\documentclass[runningheads]{llncs}
\usepackage[T1]{fontenc}
\usepackage{graphicx}
\usepackage{amsmath}
\usepackage{amssymb}
\usepackage{multirow}
\usepackage{float}
\usepackage[acronym]{glossaries}
\usepackage{algorithm} 
\usepackage{algpseudocode}

\usepackage{xcolor}

\usepackage{bm}
\begin{document}
\title{Fast and Interpretable Autoregressive Estimation with Neural Network Backpropagation}

\titlerunning{NN Backpropagation for AR Estimation}
%
\author{Anaísa Lucena\inst{1}\orcidID{0000-0002-4060-1690} \and Ana Martins\inst{1}\orcidID{0000-0003-4860-7795}
Armando J. Pinho\inst{1}\orcidID{0000-0002-9164-0016} \and
Sónia Gouveia \inst{1}\orcidID{0000-0002-0375-7610}}
\authorrunning{A. Lucena et al.}
%
\institute{IEETA/DETI/LASI, University of Aveiro, 3810-193 Aveiro, Portugal \\
\email{\{anaisalucena, a.r.martins, ap, sonia.gouveia\}@ua.pt}}
\maketitle              
\begin{abstract}

Autoregressive (AR) models remain widely used in time series analysis due to their interpretability, but convencional parameter estimation methods can be computationally expensive and prone to convergence issues. This paper proposes a Neural Network (NN) formulation of AR estimation by embedding the autoregressive structure directly into a feedforward NN, enabling coefficient estimation through backpropagation while preserving interpretability.

Simulation experiments on 125{,}000 synthetic AR($p$) time series with short-term dependence ($1 \leq p \leq 5$) show that the proposed NN-based method consistently recovers model coefficients for all series, while Conditional Maximum Likelihood (CML) fails to converge in approximately 55\% of cases. When both methods converge, estimation accuracy is comparable with negligible differences in relative error, $R^2$ and, perplexity/likelihood. However, when CML fails, the NN-based approach still provides reliable estimates. In all cases, the NN estimator achieves substantial computational gains, reaching a median speedup of $12.6\times$ and up to $34.2\times$ for higher model orders. Overall, results demonstrate that gradient-descent NN optimization can provide a fast and efficient alternative for interpretable AR parameter estimation.

\end{abstract}

\section{Introduction}

The class of \newacronym{arma}{ARMA}{Autoregressive Moving Average} \acrlong{arma} (\acrshort{arma}) and their extensions play a central role in \newacronym{ts}{TS}{Time Series}\acrlong{ts} (\acrshort{ts}) analysis. 
These models are very popular since they offer a framework to model the linear relationships of the serial dependence, while keeping the model interpretable \cite{box2015time}. Nevertheless, the simplicity offered by these models also has its own drawbacks, such as the inability to capture non-linear patterns. Moreover, the model parameters are typically estimated using Yule-Walker (YW) equations, Conditional Least Squares (CLS), and Conditional Maximum Likelihood (CML). The YW equations, are based on the method of moments, and are mainly used as a starting point for estimation via CLS and CML. The CML is the preferred estimation approach due to its asymptotically normal properties, but it is known for becoming computationally expensive as the order of the models increases, and can further suffer from convergency issues.

In contrast, \newacronym{nn}{NN}{Neural Networks} \acrfull{nn}, especially recurrent architectures \cite{tokgoz2018rnn,chang2020lstm,lin2022time}, have considerably advanced modeling and forecasting capabilities of time series by capturing complex non-linear patterns and long-term dependencies through backpropagation-based learning \cite{bengio2017deep}. However, these models often function as black boxes, making it difficult to interpret the rationale behind predictions \cite{triebe2019ar}. Moreover, their numerous parameters and deep architectures result in high computational demands, which require substantial resources \cite{ali2024comparative}. 

Recognizing the complementary strengths of these approaches, several studies explored frameworks that integrate neural networks with autoregressive structures. For instance, Tian et al. (1997) \cite{tian1997ar} introduced a feedback recurrent NN for autoregressive parameter estimation that converges to the YW solution. Hwarng et al. (2001) \cite{hwarng2001simple} showed that a simple two-layer neural network (without hidden layers) can effectively forecast linear time series generated by a wide range of ARMA$(p, q)$ processes. Triebe et al. (2019) \cite{triebe2019ar} introduced the AR-Net, a feedforward neural architecture that learns autoregressive coefficients equivalent to least-squares estimation. More recently, \acrshort{nn} approaches for time-varying AR models and bias correction in estimation have been explored \cite{jia2025time,jiang2025artificial}.

Despite these advances, most approaches focus on improving forecasting performance rather than reformulating classical parameter estimation procedures. In particular, little attention has been given to the use of neural optimization to directly estimate AR parameters while preserving the statistical structure of the model.
This work proposes a neural architecture that embeds the exact AR structure into a neural network, allowing the model coefficients to be obtained from the network weights. Moreover, the weights are constrained to ensure stationarity, which has not yet been addressed. 
Consequently, the proposed method should be interpreted as a NN-based estimator for statistical AR models rather than as a NN forecasting architecture. This distinction allows the approach to retain the interpretability and theoretical structure of AR processes, while addressing computational limitations and convergence issues observed in conventional estimation methods such as CML.

The remaining of the paper is outlined as follows. Section 2 provides background information on AR models and their estimation via conventional procedures and the novel NN-proposed framework. Then, section 3 presents the simulation study design and the performance evaluation. In section 4, the results are presented and discussed. Lastly, section 5 is devoted to the main conclusions.

\section{Methods}

This section presents the statistical formulation of AR($p$) models, the classical CML estimator and the proposed NN-based estimator for the coefficients of the model.


\subsection{Statistical formulation of AR models}

The AR model is additive and is defined as \cite{box2015time}
\begin{equation}
\label{eq:ar}
    X_t =  \sum_{i=1}^p \alpha_i  X_{t-i} + \varepsilon_t, \quad t \in \mathbb{Z}
\end{equation}
where $X_t \in \mathbb{R}$ is a continuous random variable, $\alpha_i$ with $i = 1, \dots, p$ are the model parameters, and $\varepsilon_t \sim \mathcal{N}(0, \sigma_\varepsilon^2)$ is the Gaussian innovation term, which captures the random component of the process.


For an AR($p$) process to be stationary, the coefficients $\{\alpha_1, \dots, \alpha_p\}$ must satisfy certain conditions. Rewriting \eqref{eq:ar} with the backshift operator $B$, such that $B^i X_t = X_{t-i}$, and solving for $\varepsilon_t$,

\begin{equation}
   \varepsilon_t = (1-\sum_{i=1}^p \alpha_iB^i) X_t,
\end{equation}
where $P(\alpha,B) = (1-\sum_{i=1}^p \alpha_iB^i) = \prod_{j=1}^p(1-r_j^{-1}B)$, considering $\{r_1, \dots, r_p\}$ the roots of the polynomial $P(\alpha,B) = 0$, which is the characteristic equation of AR processes. As such, an AR process can be represented as
\begin{equation}
   X_t = P(\alpha,B)^{-1} \varepsilon_t = \prod_{i=1}^p \frac{1}{1-r_i^{-1}B} \cdot \varepsilon_t.
\end{equation}
So, 
for $P(\alpha,B)^{-1}$ to be a convergent series for $|B|\leq1$, it is necessary that $|r_i^{-1}| < 1, i=1, \dots, p$. Equivalently, $|r_i|>1$, meaning that the roots of the characteristic equation must lie outside the unit circle \cite{box2015time}.

The autocorrelation (acf) and partial autocorrelation (pacf) functions are important second-order moments that describe the serial dependence of AR processes. At a given lag $k$, the acf is defined as \cite{box2015time}
\begin{equation}
    \rho_k = \frac{\operatorname{cov}(X_{t+k},X_t)}{\sqrt{\operatorname{var}(X_{t+k})\operatorname{var}(X_t)}}
    \label{eq:acf}
\end{equation}
where $\operatorname{cov}(X_{t+k},X_t)$ is the autocovariance between $X_{t+k}$ and $X_t$. Moreover, for a stationary AR($p$) process the denominator simplifies to $\sigma_\varepsilon^2$. Thus, by replacing in  \eqref{eq:acf} with the usual covariance and $\sigma_\varepsilon^2$ estimatores, $\rho_k$ can be computed. Additionally, the pacf $s_k$ is defined as
\begin{equation}
    s_k = \frac{\operatorname{cov}(X_{t+k},X_t| X_{t-1}, \dots, X_{t+k-1})}{\sqrt{\operatorname{var}(X_{t+k}| X_{t-1}, \dots, X_{t+k-1})\operatorname{var}(X_t| X_{t-1}, \dots, X_{t+k-1})}}
    \label{eq:pacf}
\end{equation}
where $|$ is the conditional operator. The pacf can be obtained from the Yule-Walker equations, which are efficiently estimated through the Durbin-Levinson (DL) recursion:

\begin{equation}
\begin{aligned}
\alpha_{1,1} &= s_1, \\
\alpha_{k,k} &= s_k, \\
\alpha_{k,j} &= \alpha_{k-1,j} - \alpha_{k,k} \,\alpha_{k-1,k-j}, \quad j = 1, \dots, k-1,
\end{aligned}
\label{eq:dl-system}
\end{equation}
where $\alpha_{k,j}, j=1,\dots,k$ are the AR coefficients of order $k$ \cite{box2015time}.

%

\subsection{AR parameter estimation via CML}

For a time series $\{x_1, x_2, \dots, x_t \}$, the conditional likelihood can be expressed as 
 \begin{equation}
  \label{eq:likelihood}
    \mathcal{L}(\theta) \propto \prod_{t=p+1}^T f_{X_t | X_{t-1}, \ldots, X_{t-p}}(x_t | x_{t-1}, \ldots, x_{t-p}; \theta),
\end{equation}
 where $f(.)$ represents the \newacronym{pdf}{pdf}{probability density function}\acrlong{pdf} (\acrshort{pdf}) of $X_t$, and $\theta$ the model parameters \cite{white2015optimal}.  
 Given that \( \varepsilon_t \sim \mathcal{N}(0, \sigma_\varepsilon^2) \), the conditional pdf of AR($p$) process is \cite{box2015time,hamilton1994time}:
\begin{equation}
    f_{X_t | X_{t-1}, \ldots, X_{t-p}}(x_t | x_{t-1}, \ldots, x_{t-p}; \theta) 
    = \frac{1}{\sqrt{2 \pi \sigma_\varepsilon^2}} 
    \exp\left(
        -\frac{(x_t - \hat{x_t})^2}{2\sigma_\varepsilon^2}
    \right),
\end{equation}
where $\hat{x_t} = \sum_{i=1}^p \alpha_i x_{t-i}$ and $(X_t | X_{t-1}=x_{t-1}, \ldots, X_{t-p}=x_{t-p}) \sim \mathcal{N}\left(\hat{x_t}, \sigma_\varepsilon^2\right)$.
 
 The goal is to find the set of parameters that maximize $\mathcal{L}(\theta)$, however the use of the log-likelihood function $\ell(\theta) = \operatorname{log}(\mathcal{L}(\theta))$ is preferred since the product in \eqref{eq:likelihood} is transformed into a summation, which makes calculations easier. This is straightforward because the maxima is attained at the same parameter values both for the likelihood and log-likelihood functions. The log-likelihood function for a Gaussian AR($p$) process with sample size $T$ is

\begin{equation}
\label{eq:likelihood_ar}
\ell(\theta) \propto - \sum_{t=p+1}^{T} \left(x_t - \hat{x_t}\right)^2.
\end{equation}

Nevertheless, it is often not possible to calculate the maximum of the log-likelihood  analytically because the function can be highly complex, involving nonlinear dependencies, and requiring summations or integrations over large datasets or distributions. Thus, numerical methods, such as gradient-based optimization, are employed to estimate the parameters of the optimization problem
\begin{equation}
    \hat{\theta} = \arg\max_{\theta} \ell(\theta),
\end{equation}
where $\hat{\theta}$ represents the CML estimate of the model parameters $\theta$.


\subsection{NN-based AR parameter estimation}

This work adopts Feedforward Neural Networks (FNNs) due to their simplicity and interpretability. This architecture allows classical AR estimation to be embedded directly into the network and optimized through backpropagation, providing a controlled setting to evaluate optimization techniques without the additional complexity of recurrent or convolutional models. Note that this simple feedforward network does not explicitly model the temporal dependencies between the lagged inputs.

\begin{figure}[h]
    \centering
    \includegraphics[width=0.77\linewidth]{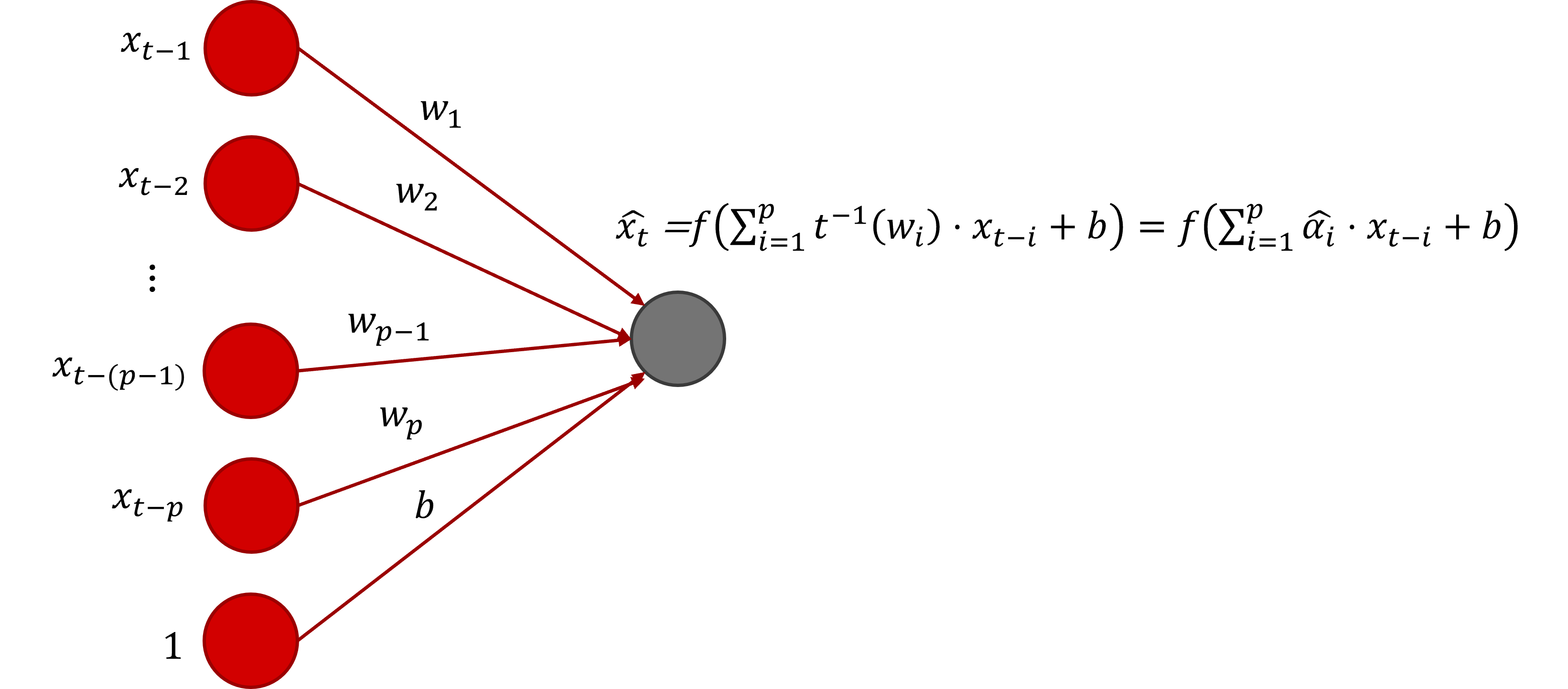}
    \caption{Feedforward neural network representation for AR($p$) time series prediction using lagged inputs $(x_{t-1},\dots,x_{t-p})$. The inverse transformation $t^{-1}(\cdot)$ maps the network weights to AR coefficients while ensuring stationarity.}
    \label{fig:NN}
\end{figure}

For time series prediction, a \acrshort{nn} can use lagged observations $(x_{t-1},\dots,x_{t-p})$ as inputs to produce a one-step-ahead prediction $\hat{x}_t$ \cite{Raza20151352}. When the order $p$ is known, input–target pairs can be constructed using a sliding window, where each input vector $(X_{t-1},\dots,X_{t-p})$ predicts the target $X_t$ \cite{suresh2020forecasting}. Fig.~\ref{fig:NN} shows a simple architecture for AR($p$) prediction with $p$ input nodes and a single output node. The network weights are then transformed into AR coefficients through $t^{-1}(\cdot)$, ensuring that the estimated parameters satisfy the stationarity constraints.

Backpropagation is the algorithm used to train \acrshort{nn}s, by minimizing the cost function, which measures the difference between predicted and actual values by adjusting the weights $w$ and bias $b$ of the network. A commonly used cost function is the \newacronym{mse}{MSE}{Mean Squared Error}\acrfull{mse} \cite{bengio2017deep}, which is the average of the squared differences between predicted and true values
\begin{equation}
\label{eq:MSE}
    \text{\acrshort{mse}} = \frac{1}{T-p-1} \sum_{t=p+1}^T (x_t - \hat{x}_t)^2.
\end{equation}
Thus, it is clear that the solution resulting from minimizing the \acrshort{mse} is equivalent to the one maximizing the log-likelihood \eqref{eq:likelihood_ar}, meaning minimizing $-\ell(\theta)$, also called \emph{perplexity}. 
\acrshort{nn}s are predominantly trained using gradient descent, which is involved  in the backward pass during training training. Thus, the gradient of the cost $C$ with respect to each parameter is computed, followed by the propagation of these errors through the network
\begin{equation}
\label{eq:backpropagation_gradient}
   \frac{\partial C}{\partial \theta_i} = \frac{\partial C}{\partial \hat{X}_t} \cdot \frac{\partial \hat{X}_t}{\partial \theta_i}.
\end{equation}
Then, the parameters are updated using a learning rate $\eta$
   \begin{equation}
   \label{eq:update_gd}
       \theta_i \leftarrow \theta_i - \eta \frac{\partial C}{\partial \theta_i}.
   \end{equation}



\section{Study design and performance evaluation}

This section describes the simulation setting, implementation details and performance metrics used to compare the NN and CML estimators.



\subsection{Simulation setting}


The CML and NN estimation approaches were trained on synthetic data generated from AR($p$)  processes. However, sampling AR coefficients in the stationarity region becomes unfeasible for higher orders. For example, a stationary AR($1$) is obtained if $\alpha_1 \in (-1,1)$, while for AR(2) it corresponds to the triangle formed by $\alpha_2+\alpha_1<1$, $\alpha_2-\alpha_1<1$, and $|\alpha_2|<1$. Thus, the complexity quickly increases with $p$ \cite{box2015time}. This is addressed by taking advantage of the correspondence between the process parameters and its pacf, 
computed through the DL recursion \eqref{eq:dl-system}, which has
much simpler stationarity constraints: $|s_i| < 1$, $i = 1, \dots, p$ \cite{jones1987randomly}.

Following \cite{jones1987randomly} (Algorithm 1), stationary AR processes can be generated by sampling the pacf from appropriate Beta distributions and converting them into AR coefficients using the DL recursion. This process ensures that the AR coefficients are uniformly distributed on the stationary region.

\begin{algorithm}[H]
\caption{AR Process Generation Algorithm}
\begin{algorithmic}[1]
\State Generate $s_1, \dots, s_p$ independently following $s_k \sim \text{Beta}\left((k+1)/2, k/2+1 \right)$
\State Use DL \eqref{eq:dl-system} to obtain $\alpha_{1}, \dots, \alpha_{p} \equiv \alpha_{p,1}, \dots, \alpha_{p,p}$
\end{algorithmic}
\end{algorithm}

The time series were generated using Gaussian innovations with fixed mean ($\mu_\varepsilon = 0$) and variance ($\sigma_\varepsilon^2 = 1$), 
and  
$1,500$ data points each. However, to reduce dependency on initial values, 
the first 500 points of each series were discarded as a burn-in period, resulting in time series of length $T=1,000$. In total, 500 processes for orders $p=1,\dots, 5$ with $50$ repetitions each were generated, resulting in $125,000$ time series.

\subsection{Implementation details}


Initial conditions for both approaches were obtained using the YW equations.
In the NN, the lagged observations \((x_{t-1}, x_{t-2}, \ldots, x_{t-p})\) were fed as input vectors using a sliding window approach, and the network produced a one-step-ahead prediction (\(\hat{x}_t\)) at the output layer. Since the intercept term of the time series can be calculated using its mean and the estimated AR coefficients, the bias was deemed redundant and, as such, $b=0$. To maintain linearity in the AR framework, the identity function was used as the activation function at the output. Moreover, the weights of the network, $w$, are a function of the AR coefficients of the model, such that $\hat{\alpha_i} \equiv t^{-1}(w_i)$. 
Thus, the transformations made to the coefficients and, reversely, to the weights were
\begin{equation}
\left\{
  \begin{array}{ll}
    t(\alpha) = \text{arctanh}(\text{DL}^{-1}(\alpha)) \\
    t^{-1}(w) = \text{DL}(\tanh(w))
  \end{array}.
\right.
\end{equation}
The hyperbolic tangent ensures that the weights are always $|\tanh(w)|<1$, and the DL recursion converts that result to an estimate of the coefficients $\hat{\alpha}$, that always corresponds to a stationary process, while allowing the weights $w$ to be trained in an unconstrained space.

The training proceeds by minimizing the cost function, i.e., the \acrshort{mse}. 
The Adam optimizer was chosen for updating the network parameters, due to its adaptive learning rate $\eta$, enhancing both convergence speed and stability. Given the relatively small sample size of the time series, full batch training was employed. Algorithm 2 describes the procedure to obtain estimates of the AR coefficients with backpropagation. The training was performed using PyTorch 2.5.0.

\begin{algorithm}[h]
\caption{Neural Network Training Process}\label{alg:inar_nn}
\begin{algorithmic}[1]

    \State Use Yule-Walker equations to obtain initial estimates $\alpha_{YW}$
    \State Initialize neural network model with parameters $w = t(\alpha_{YW})$

    \State Initialize optimizer (e.g. Adam) and cost function $C$, as in \eqref{eq:MSE}.
    \State Split data: for AR order $p$, construct input sequences $X = \{ [x_{t-p}, \dots, x_{t-1}] \}_{t=p+1}^{T}$ and targets $Y = \{ x_t \}_{t=p}^{T}$
    \For{each epoch $e = 1, \dots, E$}
        \State Compute expected values $\hat{Y}_t$ 
        \State Compute cost $C$ 
        \State Update $w$ using gradient descent (Eq.\ref{eq:update_gd})
        \If{convergence criterion met}
            \State Break
        \EndIf
    \EndFor
\end{algorithmic}
\end{algorithm}

For CML estimation, the R package \emph{arima} \cite{R-arima2} was used. The $include.mean$ option was not selected, since a bias term is not used in the NN. 
Also, by default, $arima$ reparametrizes the AR parameters during optimization to enforce stationarity (option $transform.pars$), similarly to the weight transformations performed on the NN. The optimizer used was the BFGS (Broyden– Fletcher– Goldfarb– Shanno) algorithm. Algorithm 3 presents the CML procedure. 
\begin{algorithm}[h]
\caption{Conditional Maximum Likelihood Process}
\begin{algorithmic}[1]
\State Use Yule-Walker equations to obtain initial estimates for $\alpha$ 
\State Define the log-likelihood function as in \eqref{eq:likelihood_ar} 
\For {each epoch $e = 1, \dots, E$}
    \State Maximize the log-likelihood with an iterative optimization algorithm (e.g. BFGS)
    \State Update parameter estimates.
        \If{convergence criterion met}
            \State Break
        \EndIf
\EndFor
\end{algorithmic}
\end{algorithm}

Regarding convergence, a maximum of 10,000 epochs were allowed for each time series. However, training can stop earlier if the improvement in the cost function becomes negligible. This stopping criteria was designed to be equivalent to the \emph{reltol} parameter in R. Specifically, the training stops when $\bigl| C_t - C_{t-1} \bigr| < \Delta \, \bigl( |C_{t-1}| + \Delta \bigr)$, with  $\Delta = 10^{-6}$.

\subsection{Performance metrics}

To evaluate the proposed NN architecture in comparison with CML estimation, five criteria were considered: the coefficient relative error $ \left|(\alpha_i-\hat{\alpha}_i)/\alpha_i\right| $ for $i=1, \cdots, p$, the coefficient of determination $R^2$, the value of the cost functions at the optimum solution, the computation time and the convergence success rate of each method. The coefficient of determination $R^2$ defined as
\begin{equation}
    R^2 = 1-\frac{\sigma_\varepsilon^2}{\operatorname{var}(X_t)},
\end{equation}
where \(\sigma_\varepsilon^2\) is the variance of the white-noise innovations \(\varepsilon_t\), and \(\operatorname{var}(x_t)\) is the variance of the process \cite{nelson1976interpretation}. For a stationary AR process, this variance can be obtained from the power spectrum \cite{von1999statistical}:
\begin{equation}
   \operatorname{var}(X_t)
    =
      \sigma_\varepsilon^2 \cdot 2\int_0^{1/2}
        \frac{1}
        {\left|1 - \sum_{k=1}^p \alpha_k e^{-2\pi i k \omega}\right|^2}
      \,d\omega.
\end{equation}

In addition, the cost functions associated with both methods were evaluated for each set of estimated coefficients. For the NN, the MSE in \eqref{eq:MSE} was used, while for CML the likelihood was computed as implemented in the \emph{arima} package source code \cite{R_arima_source}. Under Gaussian innovations, minimizing the MSE is equivalent to maximizing the likelihood; therefore, for consistency, the analysis is reported in terms of MSE and perplexity differences between the two approaches. Computation time was also recorded for every estimation procedure, with all experiments run on the same machine (12th Gen Intel(R) Core(TM) i7-1255U processor with 16GB of RAM). Finally, convergence success rates were compared across methods and AR orders, since robustness of the estimation procedure is one of the main aspects under study.

\section{Results}

This section presents the comparison between the NN-based and CML estimators. First, the convergence behavior is examined, highlighting the conditions under which CML fails. Second, their estimation accuracy, cost-function values and computational efficiency are compared. Finally, the effect of increasing model order on the performance and scalability of both estimators is analyzed.



\subsection{Convergence in CML and NN estimation}

Table 1 shows the number of times that both CML and NN successfully completed the estimation procedure. The NN is able to complete the estimation procedure for all instances evaluated. In contrast, the CML struggled for a considerable number of time series, which increases from around 25\% for $p=1$ to nearly 80\% for $p=5$. The overall success rate of CML is around 55\%, with failures occuring due to a ``non-finite value supplied by optim'' error. These failures are expected when AR roots approach the unit circle, since the likelihood surface becomes ill-conditioned and difficult to optimize numerically.


\begin{table}[h]
    \centering
        \caption{Comparison of successful and failed time series evaluations, according to order $p$ for CML and NN based estimation.}
    \begin{tabular}{c|cc|cc}
     & \multicolumn{2}{c|}{\textbf{CML}} & \multicolumn{2}{c}{\textbf{NN}}\\
    $p$& $\#$Success & $\#$Failure&  $\#$Success & $\#$Failure\\
    \hline
        1 &  18760 ($75\%$) & 6240 ($25\%$) & 25000 & 0\\
         2 & 13844 ($55\%$) & 11156 ($45\%$)& 25000 & 0\\
         3 & 9745 ($39\%$) & 15255 ($51\%$) & 25000 & 0\\
         4 & 8398 ($34\%$) & 16602 ($66\%$) & 25000 & 0\\
         5 & 5484 ($22\%$) & 19516 ($78\%$) & 25000 & 0\\
         \hline
         Total & 56231 ($55\%$) & 68769 ($45\%$) & 125000 & 0
    \end{tabular}
    \label{tab:cml_status}
\end{table}

\begin{figure}[h]
\centering
\includegraphics[scale=0.33]{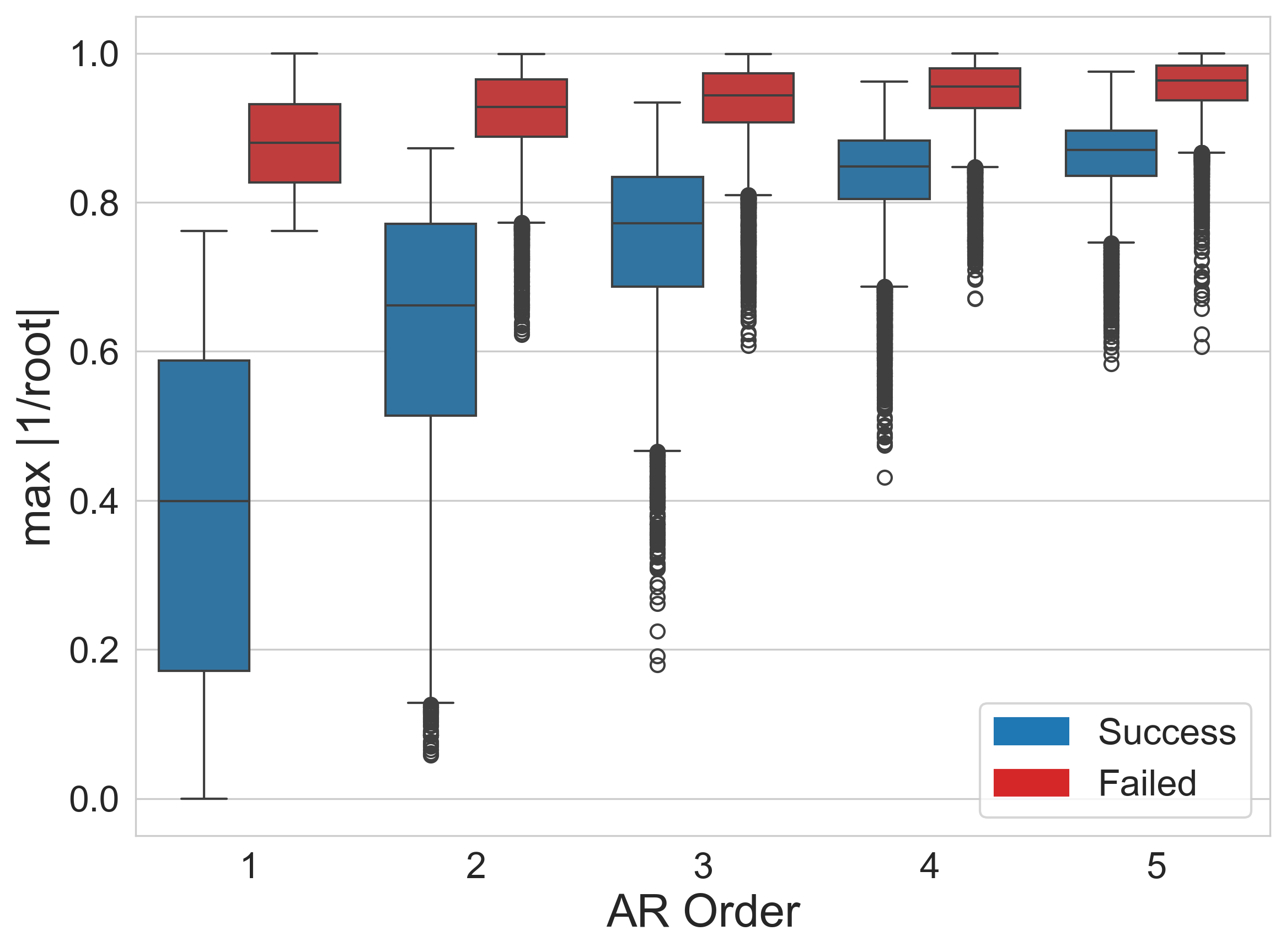}
\caption{Boxplot of the distribution of the maximum absolute inverse root (YW), by AR order and CML status. }
\label{fig:yw_failed_cml_root}
\end{figure}

Both methods start with (the same) initial values computed from the YW equations, which return stationary estimates for each time series. Fig.~\ref{fig:yw_failed_cml_root} shows the distribution of maximum absolute inverse roots of the Yule-Walker conditions for the AR($p$) models according to CML status (success or failure). It is evident that the CML fails more frequently when at least one of the roots lies close to the unit circle. For the AR($1$) process, where only a single root exists, a clear cutoff is observed at $\alpha_1 \approx 0.7616$. As $p$ increases, interactions among the roots become more complex, making it harder to identify the point for CML failure.

The convergence issues observed for CML when the roots approach the unit circle can be explained by the numerical properties of the likelihood surface. When an AR process is close to the stationarity boundary, the likelihood becomes nearly flat and the optimization problem becomes ill-conditioned, making parameter estimates highly sensitive to small perturbations in the data. In such situations, second-order optimization methods such as BFGS may struggle to approximate the curvature of the likelihood surface and can produce unstable updates or non-finite evaluations. In contrast, the proposed NN estimator relies on first-order gradient updates through backpropagation and performs optimization in an unconstrained parameter space via the Durbin–Levinson reparameterization. This produces a better-conditioned optimization problem, improving numerical stability when the process approaches the stationarity boundary.

\begin{figure}[h]
\centering
\begin{tabular}{cc}
(a) & (b) \\
\includegraphics[scale=0.3]{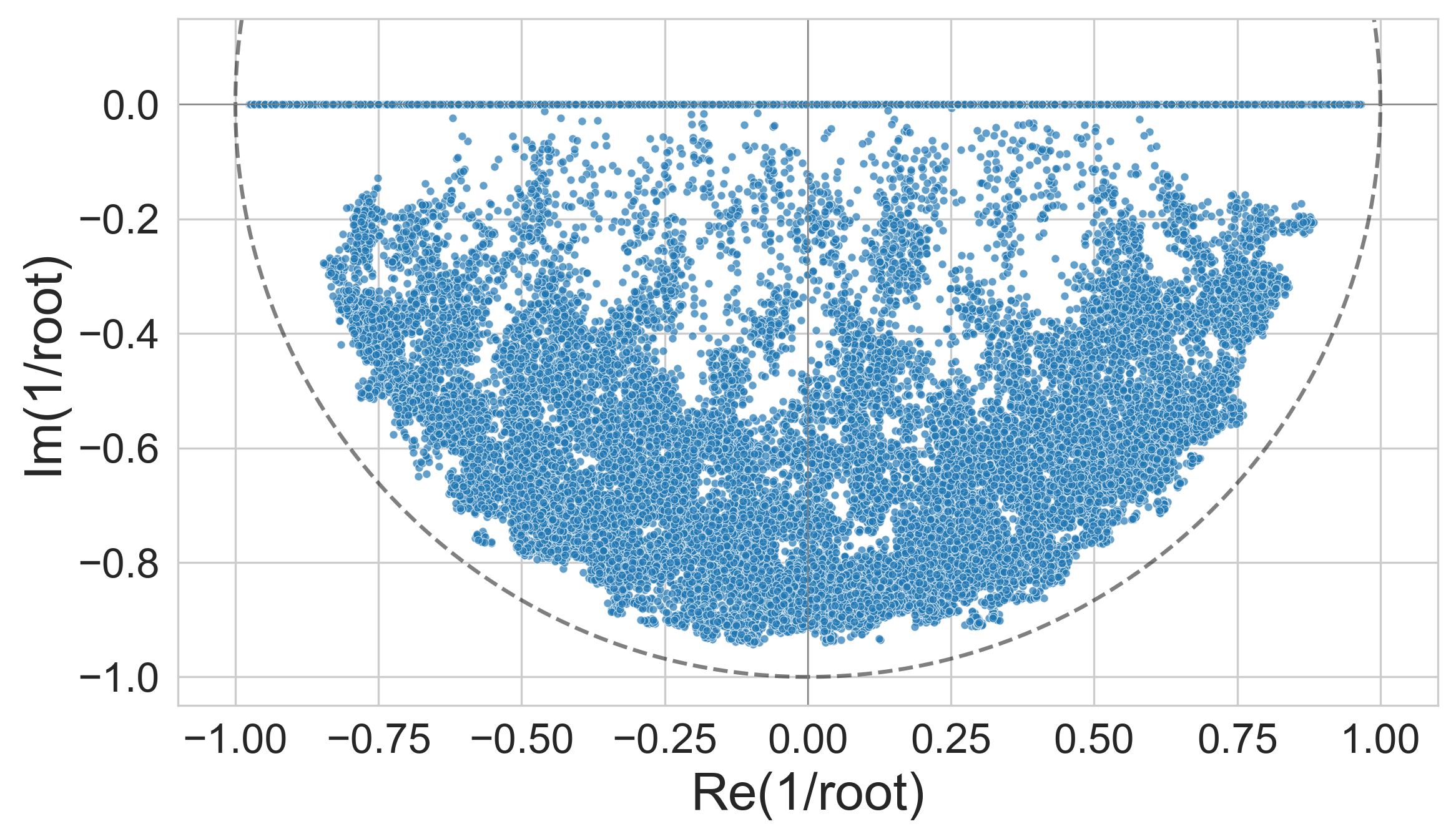}
&
\includegraphics[scale=0.3]{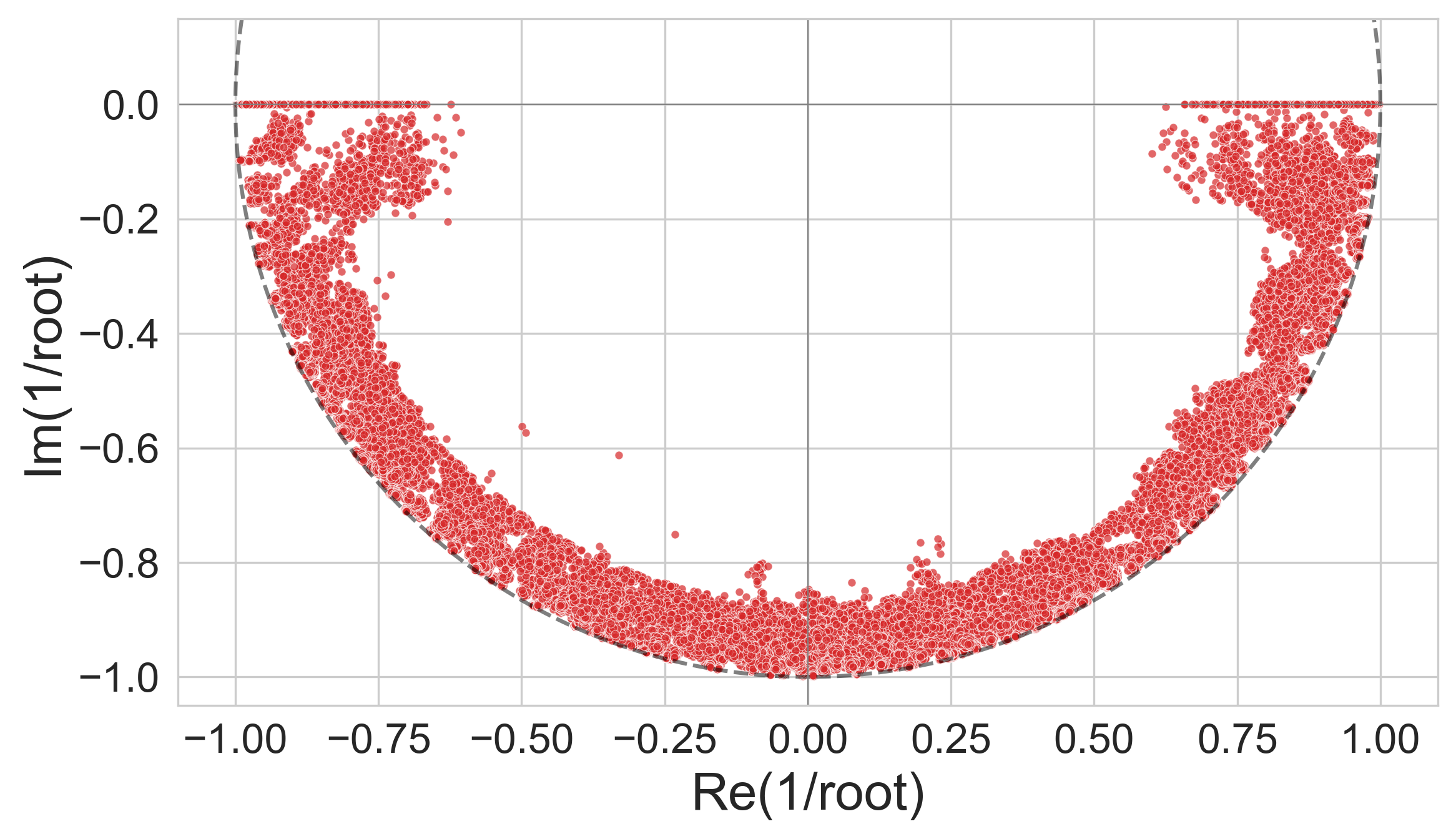} \\
(c) & (d) \\
\includegraphics[scale=0.3]{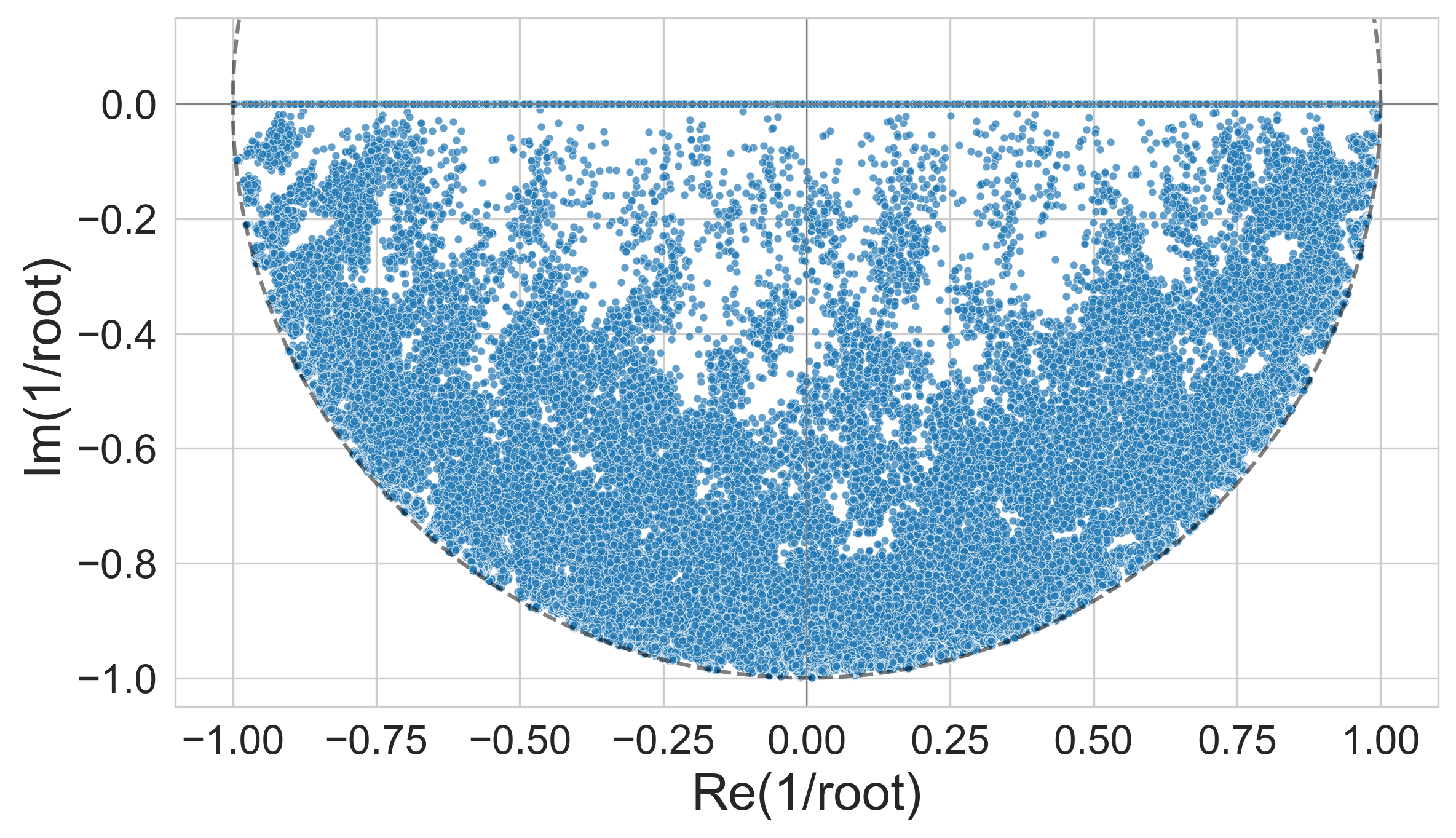}
&
\includegraphics[scale=0.3]{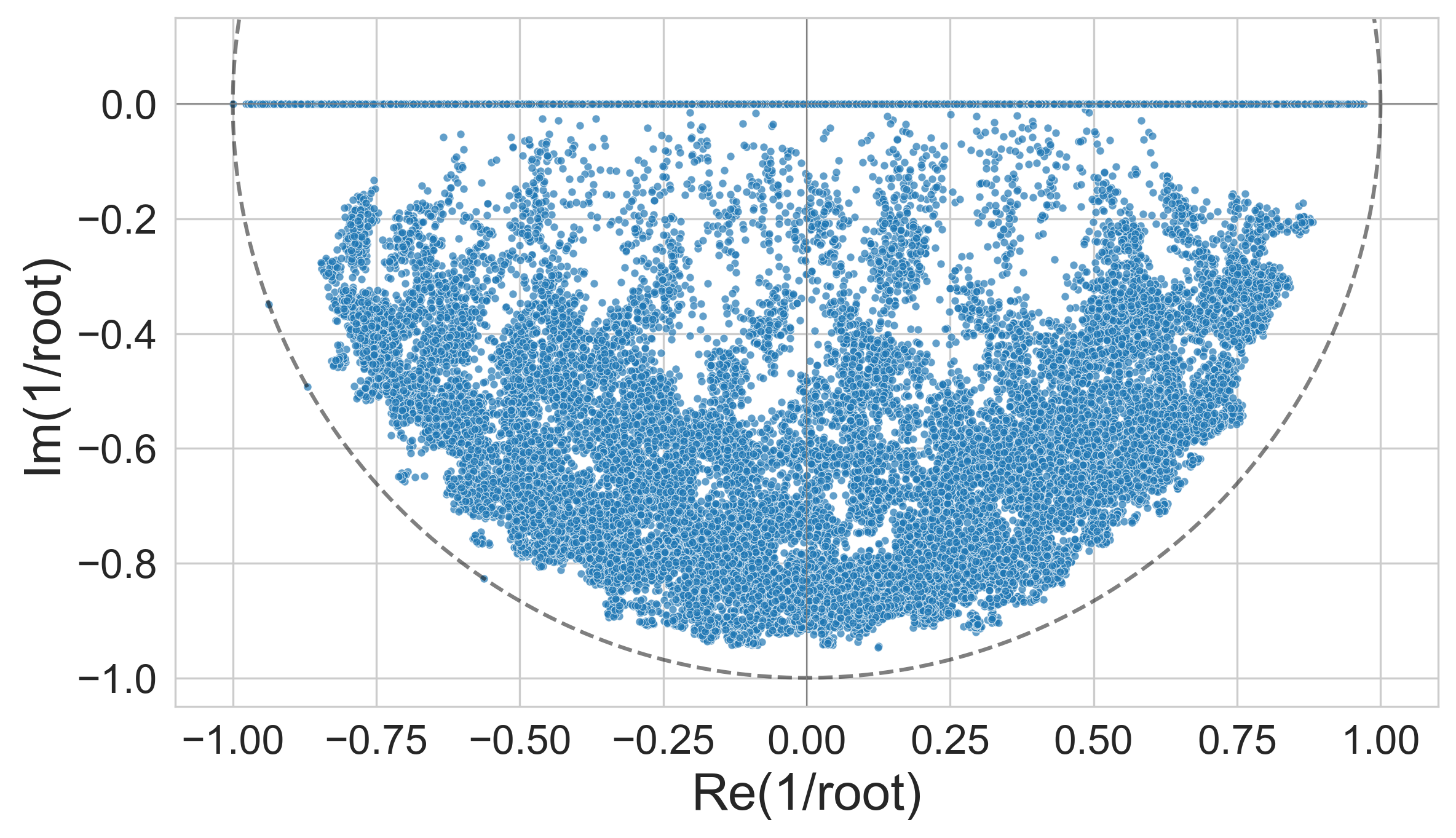} \\

\end{tabular}
\caption{Maximum absolute inverse root of the initial YW estimates within the unit circle (cropped for visualization): (a) CML successful, (b) CML unsuccessful, and coefficients estimated from (c) NN and (d) CML, when both methods converge.}
\label{fig:yw_failed_cml_root_circle}
\end{figure}

Fig.~\ref{fig:yw_failed_cml_root_circle}(a,b) further illustrates this behavior by showing the maximum absolute inverse roots of the initial coefficients obtained from the YW method, according to the CML convergence status. When CML fails (b), the roots are clearly concentrated closer to the unit circle boundary. In contrast, when CML succeeds (a), the roots tend to lie further inside the stationary region. Fig.~\ref{fig:yw_failed_cml_root_circle}(c,d) show the maximum absolute inverse roots of the estimated coefficients obtained with the NN (c) and CML (d) when both methods converge. The NN estimates span the entire stationary region, while the CML estimates appear more restricted to regions farther from the stationarity boundary. As above mentioned, this pattern is consistent with the known numerical difficulties of likelihood-based estimation near the stationarity boundary, where small changes in the coefficients can lead to large changes in the root locations, making the optimization problem poorly conditioned.


\subsection{Accuracy, precision and computational efficiency}

\begin{figure}[t]
\centering
\begin{tabular}{cc}
(a) & (b) \\
\includegraphics[scale=0.3]{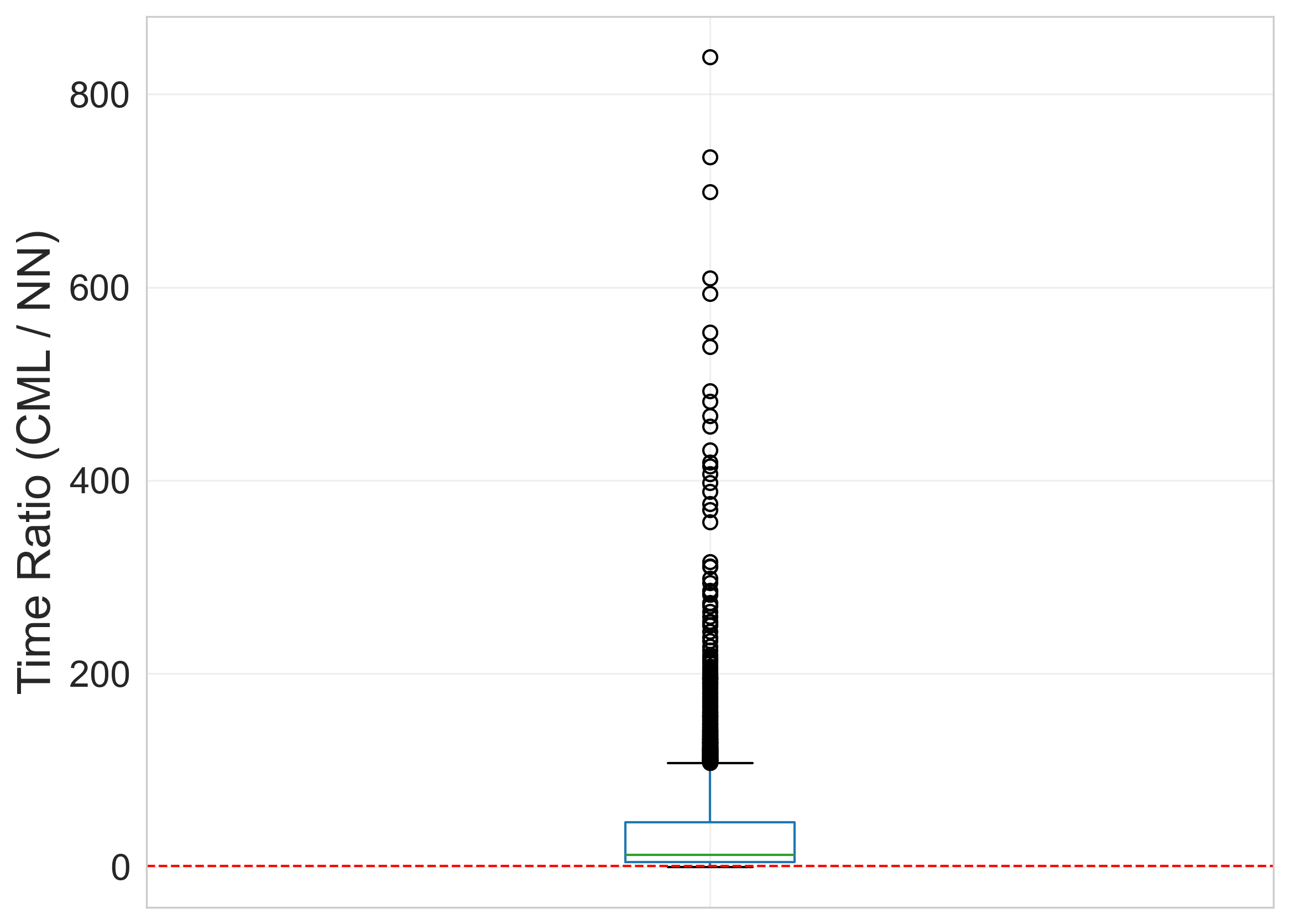} &
\includegraphics[scale=0.3]{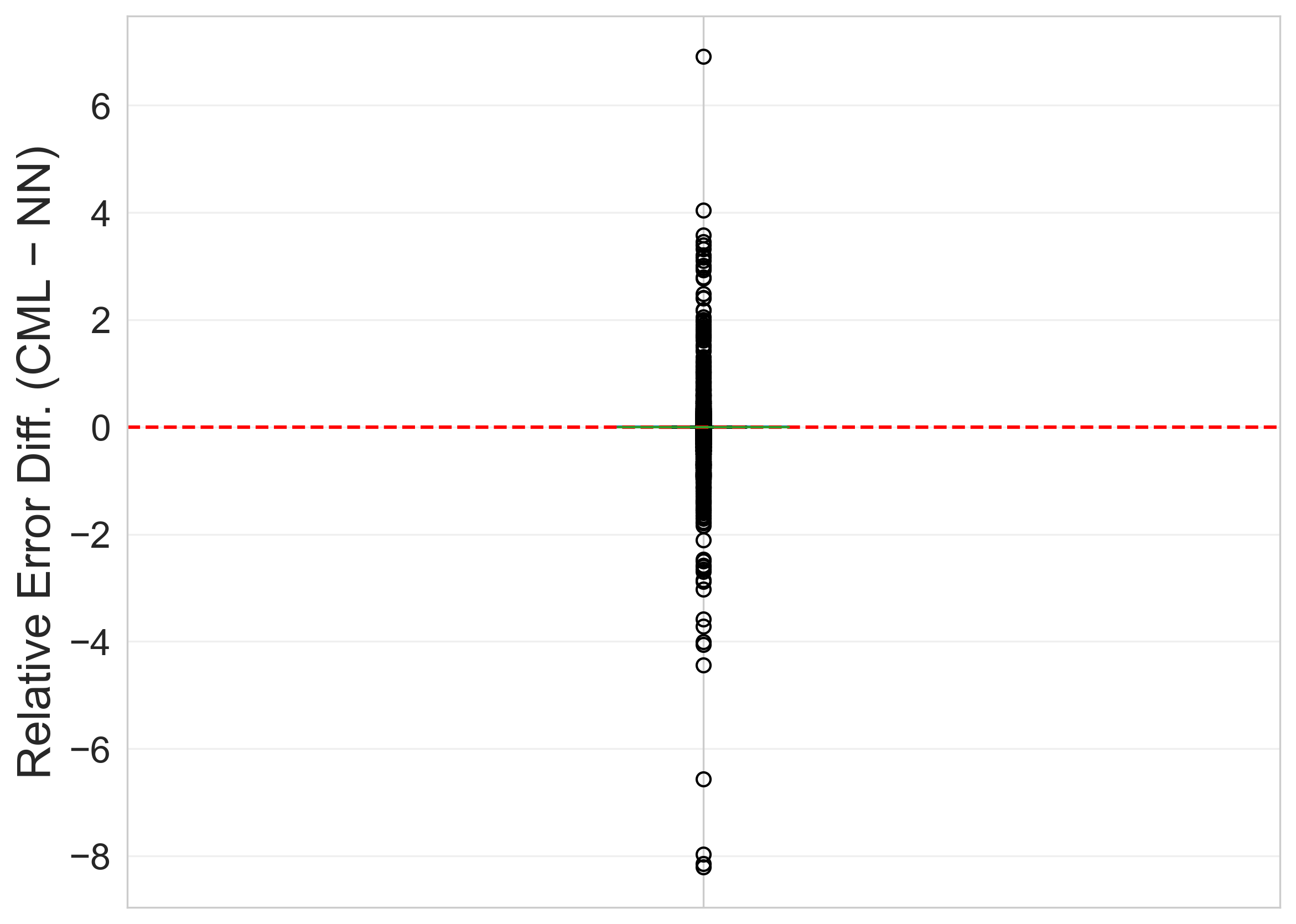} \\
(c) & (d) \\
\includegraphics[scale=0.33]{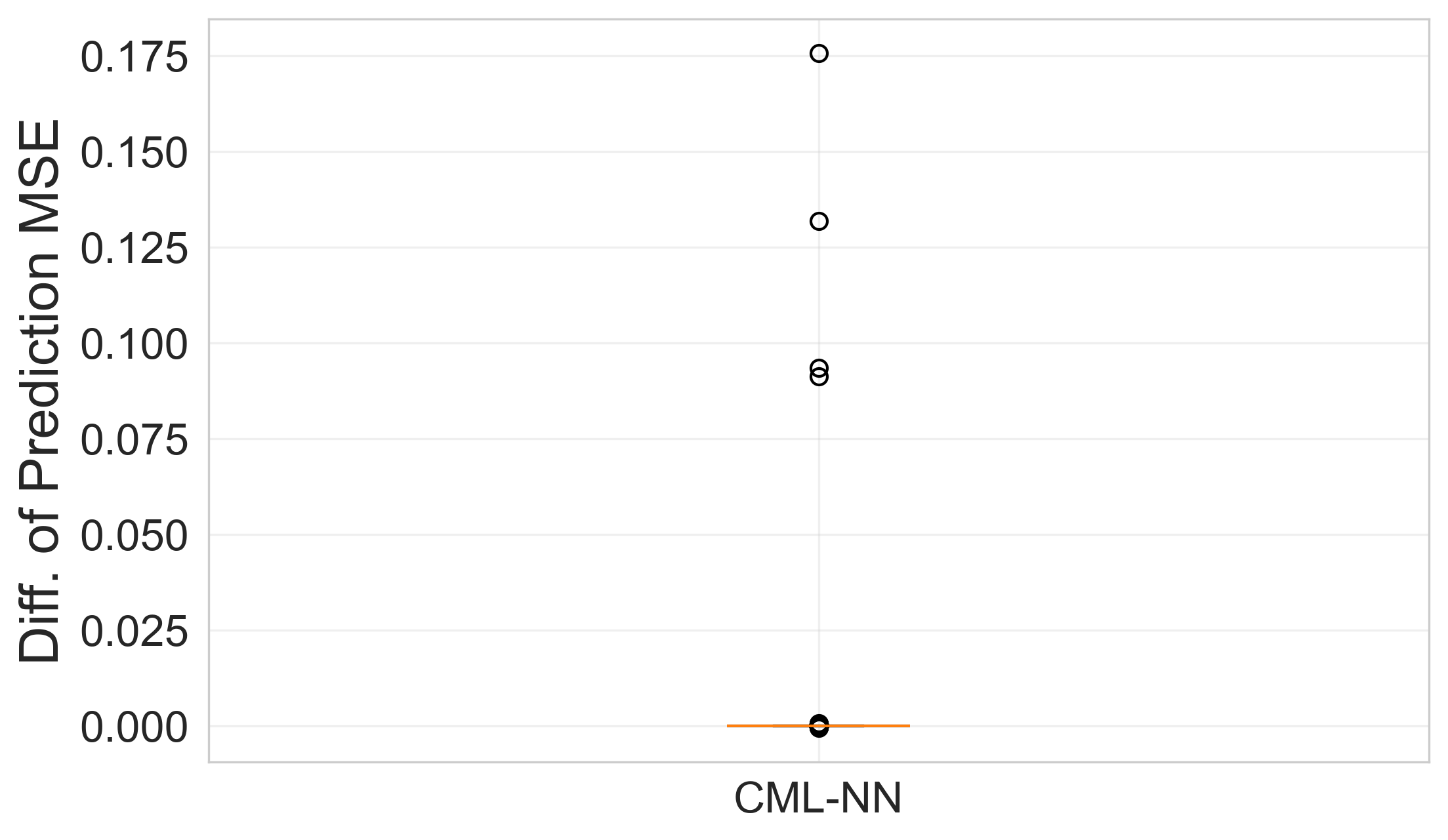} &
\includegraphics[scale=0.33]{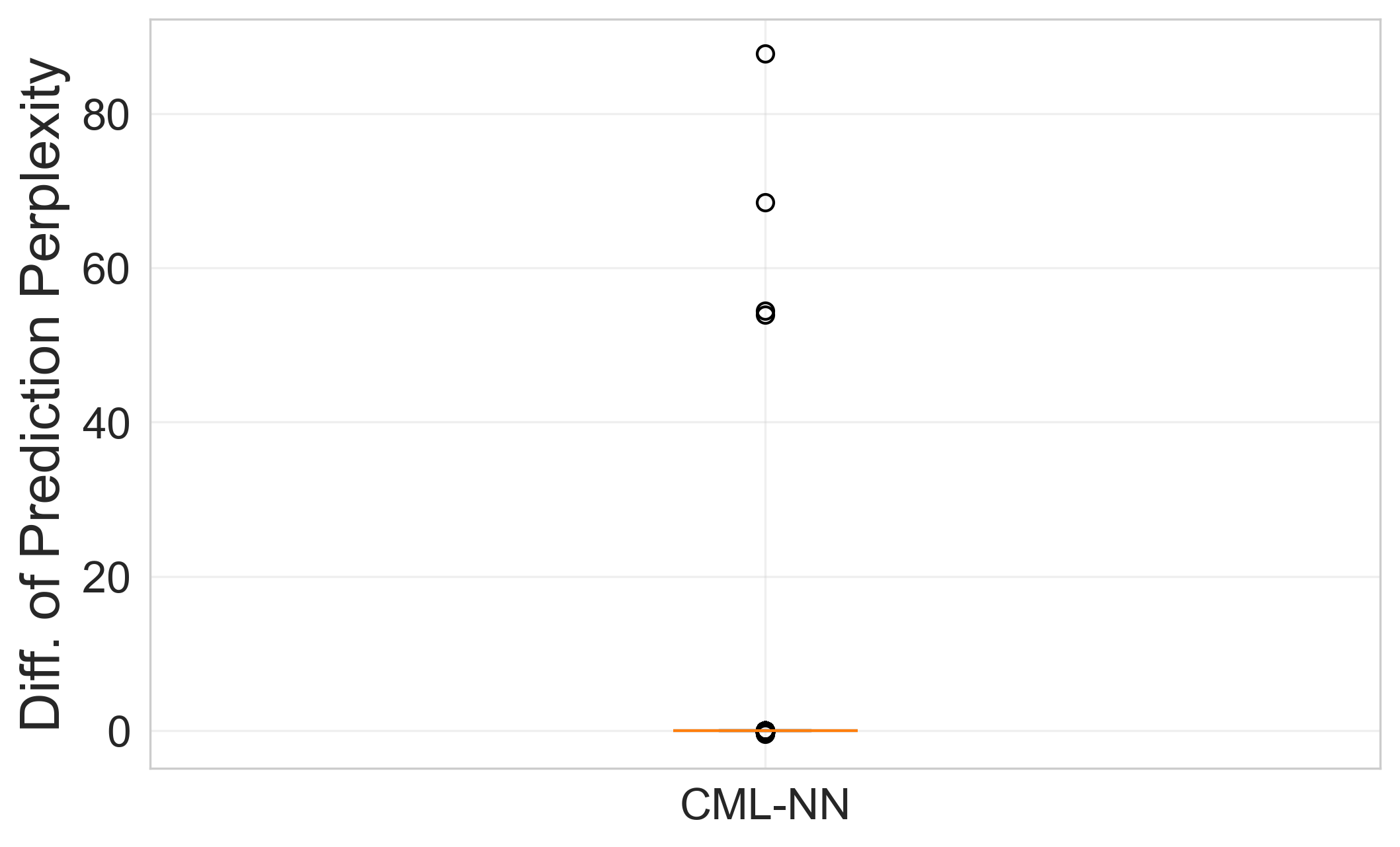} \\
\end{tabular}
\caption{Pairwise comparison between CML and NN estimation: (a) computation time ratio (CML/NN), (b) relative error difference (CML$-$NN) of the estimated coefficients, (c) MSE difference (CML$-$NN), (d) perplexity difference (CML$-$NN).}
\label{fig:cml_nn_comparison}
\end{figure}

Since the CML fails to obtain estimates for a large portion of the simulated time series, the first results of this section focus on the 55\% of cases where both CML and NN successfully converge. Fig.~\ref{fig:cml_nn_comparison}(a) shows the computation time ratio (CML/NN). The NN estimator is substantially faster, with a median speedup of 12.6, and in some cases CML can be up to 800 times slower. The relative error was computed for each estimated coefficient obtained via CML and NN, and the paired difference (CML$-$NN) was then evaluated. The results are shown in Fig.~\ref{fig:cml_nn_comparison}(b). The median relative error difference is $10^{-5}$ and the mean difference is approximately $-10^{-3}$. Overall, CML tends to produce slightly lower relative errors on average, although the NN achieves lower errors for a slightly larger number of time series.


Under Gaussian innovations, minimizing the MSE is equivalent to maximizing the likelihood. Thus, Fig.~\ref{fig:cml_nn_comparison}(c,d) compares the cost functions obtained with CML and NN. The paired differences (CML$-$NN) in MSE and perplexity have very similar distributions, with median values of $3.17\times10^{-8}$ and $-9.98\times10^{-4}$, respectively. The mean differences ($1.20\times10^{-5}$ for MSE and $9.56\times10^{-4}$ for perplexity) suggest a slight advantage for the NN on average, mainly due to a few cases where CML produces substantially worse cost values. This effect is mainly driven by a few instances where CML produces substantially worse cost values, while the NN rarely performs significantly worse than CML.

\begin{figure}[h!]
    \centering
    \includegraphics[width=0.6\linewidth]{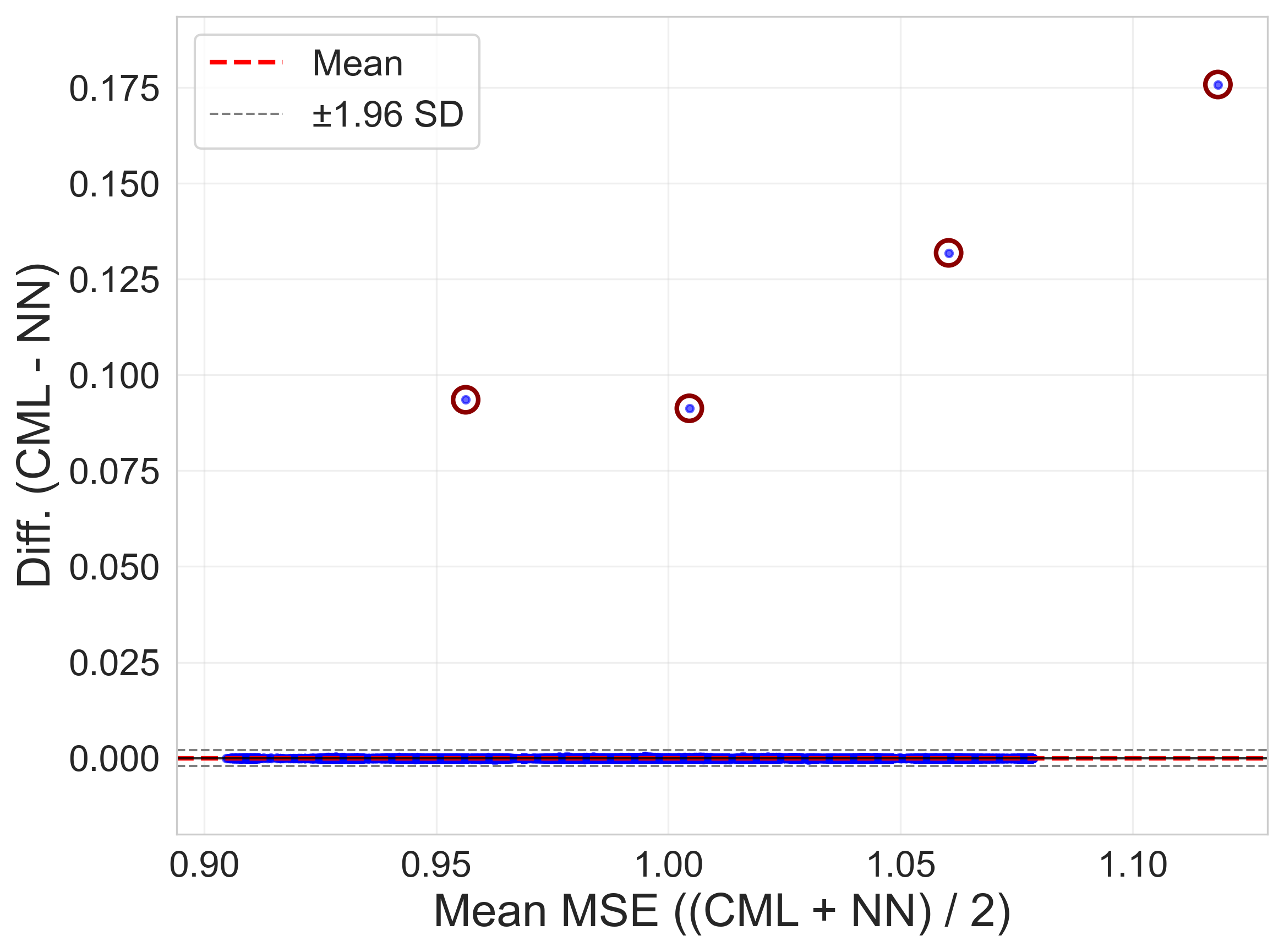}
    \caption{Bland-Altman Plot for the MSE Cost Function.}
    \label{fig:bland_altman_mse}
\end{figure}

 To better understand this discrepancy, a Bland-Altman plot  was made for the MSE cost function (Fig. \ref{fig:bland_altman_mse}). Results show that, for most time series, there is no significant difference between the cost function, since all but four points (highlighted in red circles) are within the $\pm 1.96$ standard deviation interval, i.e., a 95\% confidence interval. Results were equivalent for perplexity (data not shown). The four time series identified as outliers in the Bland–Altman analysis were examined in more detail. The results, summarized in Table~\ref{tab:4_bad_cml}, are rounded to two decimal places for readability; values reported as 1.00 correspond to 1.00010 or 1.00001 in the original estimates. In all cases, the CML procedure produced estimates that deviate substantially from both the true data-generating coefficients and the initial Yule–Walker estimates. The corresponding roots are also markedly different from those of the underlying process, with at least one root lying very close to the stationarity boundary. This behavior highlights the numerical instability of the CML estimation when the process approaches the admissible region. Consequently, these time series were excluded from the remaining analysis.
 

\begin{table}[h!]
\caption{Absolute roots and corresponding coefficients across the ``outlier'' time series. Results are rounded to the second decimal place for readability, CML values of 1.00 are, in fact, 1.00010 or 1.00001.}
\centering
\begin{tabular}{c l l l}
\hline
\textbf{TS} & \textbf{Method} & \textbf{Absolute Roots} & \textbf{Coefficients} \\
\hline

\multirow{4}{*}{1}
& Process & (1.16, 1.16) 
          & (1.52, 0.75) \\
& YW      & (1.15, 1.15) 
          & (1.53, 0.76) \\
& NN      & (1.14, 1.14) 
          & (1.54, 0.77) \\
& CML     & (1.00, 1.00) 
          & (1.74, 1.00) \\
\hline

\multirow{4}{*}{2}
& Process & (1.97, 1.23) 
          & (0.30, 0.41) \\
& YW      & (2.02, 1.24) 
          & (0.31, 0.40) \\
& NN      & (2.02, 1.24) 
          & (0.31, 0.40) \\
& CML     & (1.84, 1.00) 
          & (0.46, 0.54) \\
\hline

\multirow{4}{*}{3}
& Process & (2.17, 1.21, 1.21) 
          & (1.12, 0.04, 0.32) \\
& YW      & (2.04, 1.20, 1.20) 
          & (1.09, 0.09, 0.34) \\
& NN      & (2.04, 1.20, 1.20) 
          & (1.10, 0.08, 0.34) \\
& CML     & (1.76, 1.00, 1.00) 
          & (1.31, 0.06, 0.57) \\
\hline

\multirow{4}{*}{4}
& Process & (1.10, 1.10, 1.06, 1.06) 
          & (0.94, 1.65, 0.84, 0.74) \\
& YW      & (1.11, 1.11, 1.07, 1.07) 
          & (0.90, 1.58, 0.80, 0.72) \\
& NN      & (1.11, 1.11, 1.07, 1.07) 
          & (0.92, 1.59, 0.81, 0.72) \\
& CML     & (1.06, 1.06, 1.00, 1.00) 
          & (1.00, 1.75, 0.88, 0.89) \\
\hline
\label{tab:4_bad_cml}
\end{tabular}
\end{table}

So far, it has been shown that NN estimation can compete with classical CML estimation in terms of coefficient accuracy and proximity to cost function, with clear superior computational efficiency. Still, it is relevant to analyze if, in the instances where CML struggles to provide estimates, the NN returns quality estimates. As such, a paired analysis of the perplexity and coefficient of determination $(R^2)$ of the NN and the process coefficients is conducted.

\begin{figure}[h]
\centering
\begin{tabular}{cc}

(a) & (b) \\
\includegraphics[scale=0.3]{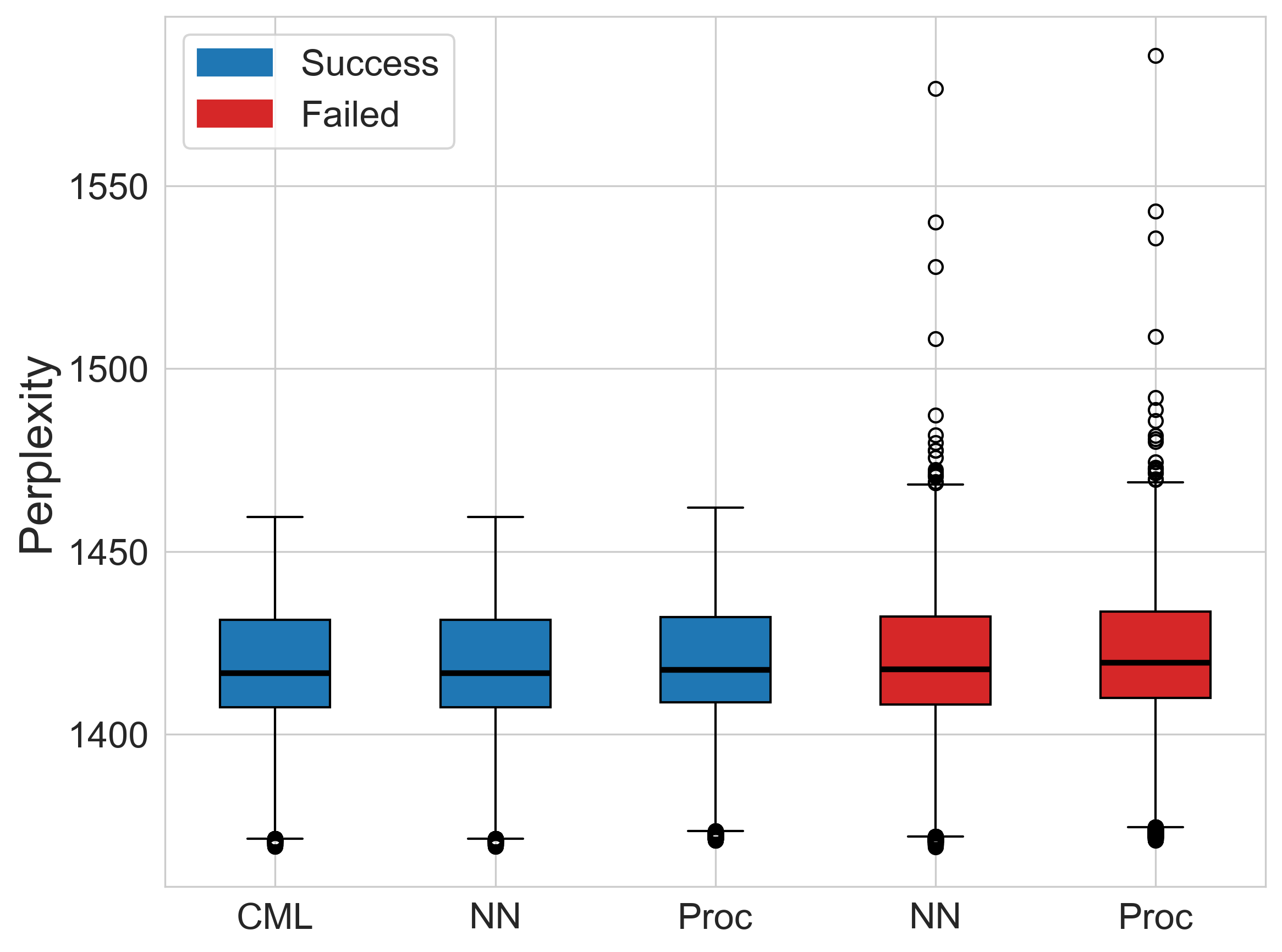}
&
\includegraphics[scale=0.3]{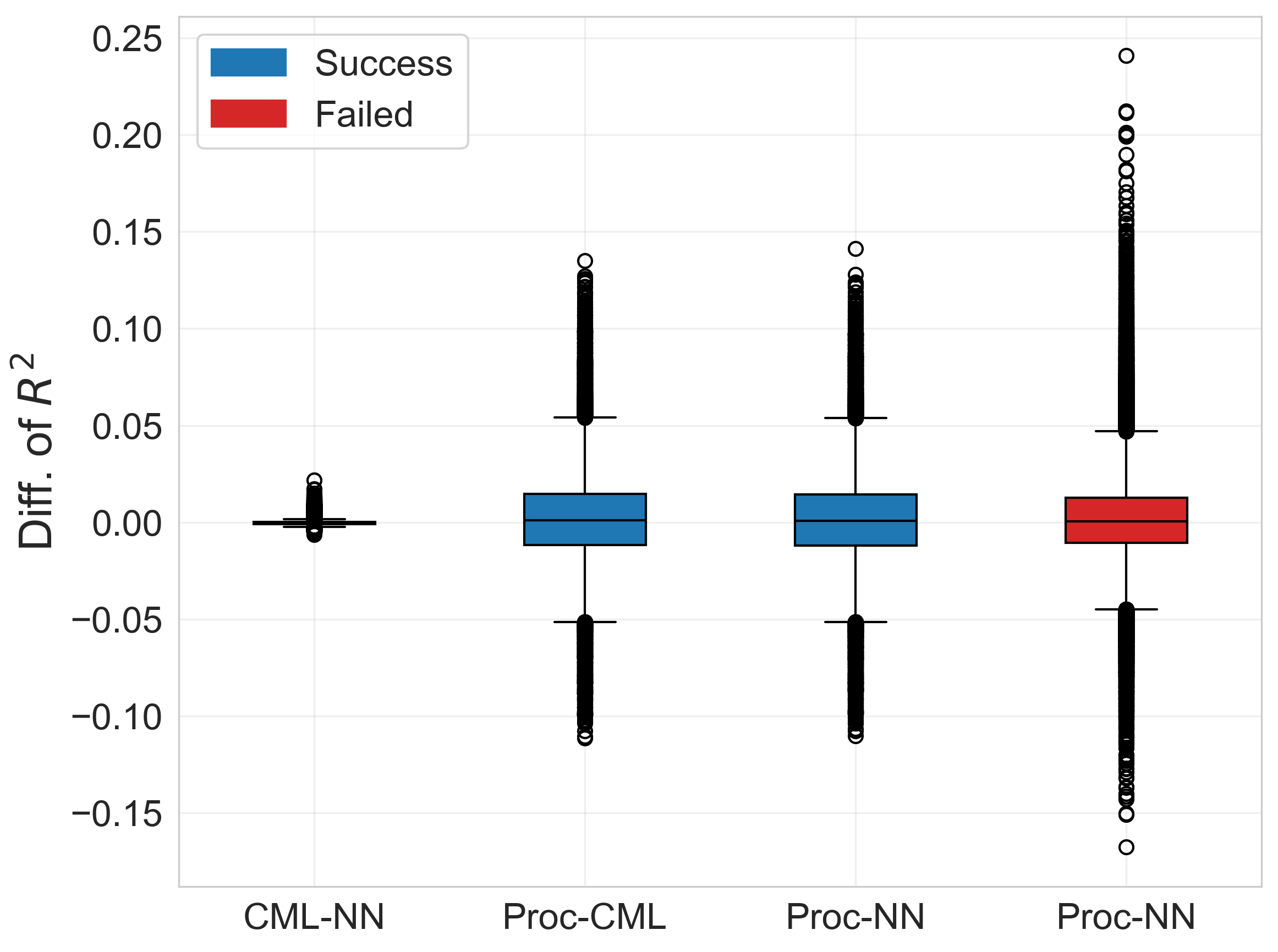} \\
\end{tabular}
\caption{Boxplot of (a) perplexity distribution and (b) coefficient of determination $(R^2)$, for CML, NN, and process coefficients (Proc), according to CML status.}
\label{fig:yw_nn_failed}
\end{figure}

Fig. \ref{fig:yw_nn_failed}(a) shows that the distribution of perplexity is similar for CML, NN and the process coefficients (Proc) when CML is successful. When CML fails, 
both the NN and process coefficients have slightly higher perplexity. However, they present a similar distribution, suggesting that the NN behavior is consistent with the process. 
Furthermore, the differences in terms of $R^2$ corroborate that CML and NN differences are negligible when CML is successful. Thus, the $R^2$ differences between the pairs Proc-CML and Proc-NN for successful CML have a similar distribution. Finally, when CML fails, the $R^2$ differences of Proc-NN have a similar median to the remaining settings, but higher variability and more outliers, according to perplexity results. Thus, the results support the ability of the NN method to consistently produce identical results to CML but, more importantly, to deliver consistent results with the process coefficients even when CML fails due to proximity to the stationarity boundary.



\subsection{Scalability with model order}

\begin{figure}[H]
\centering
\begin{tabular}{cc}
(a) & (b) \\
\includegraphics[scale=0.3]{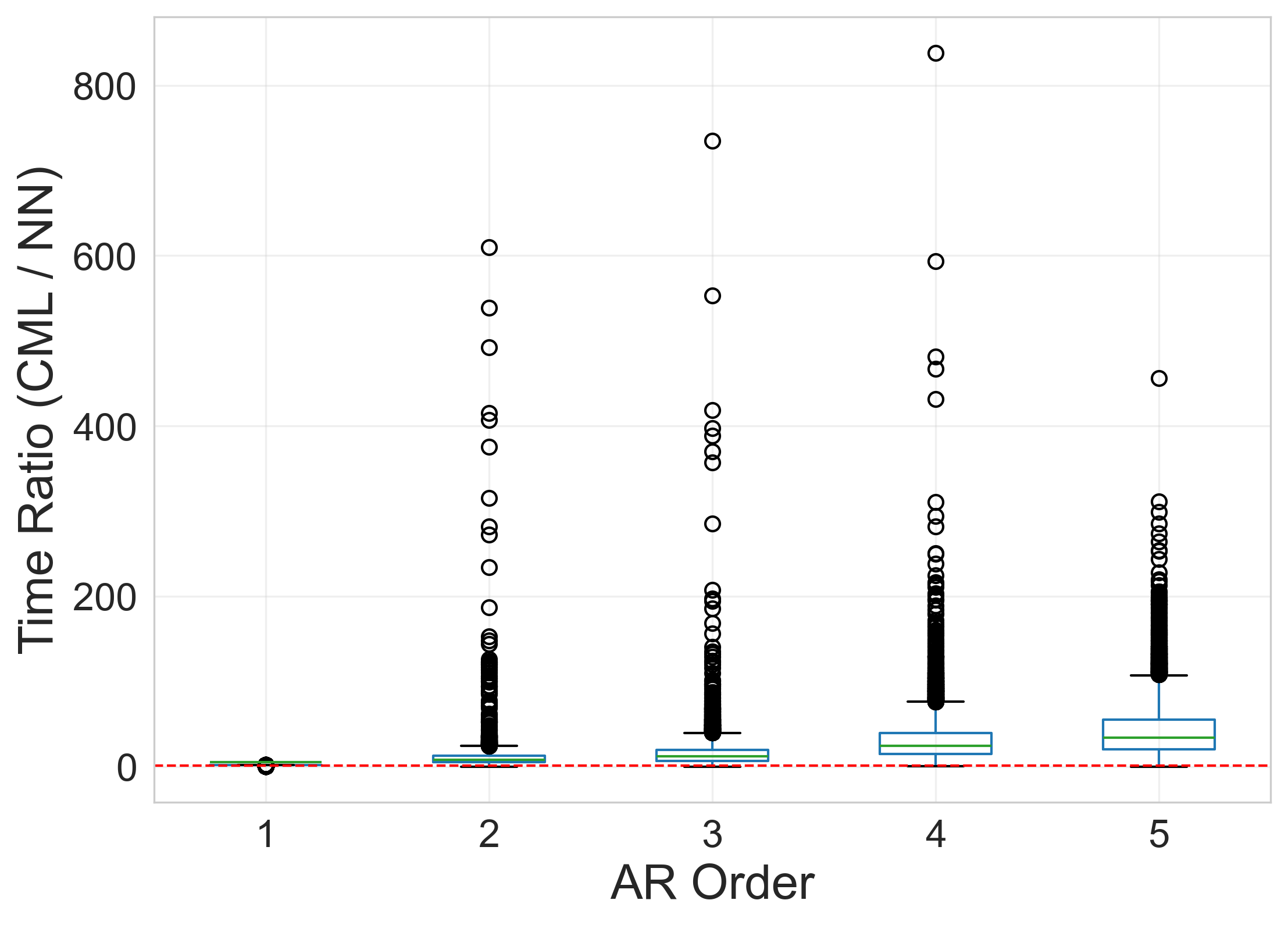} &
\includegraphics[scale=0.3]{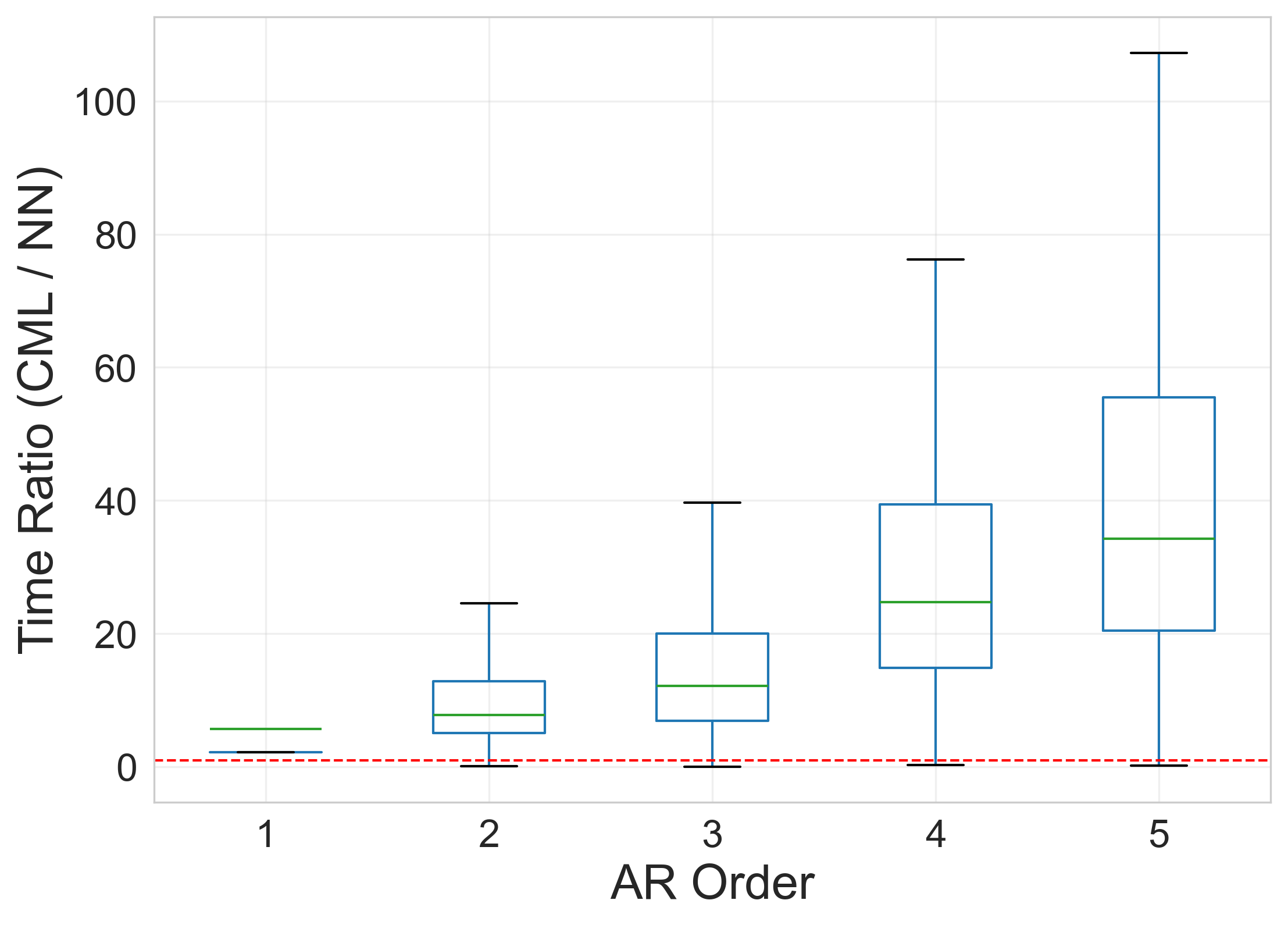} \\
(c) & (d) \\
\includegraphics[scale=0.3]{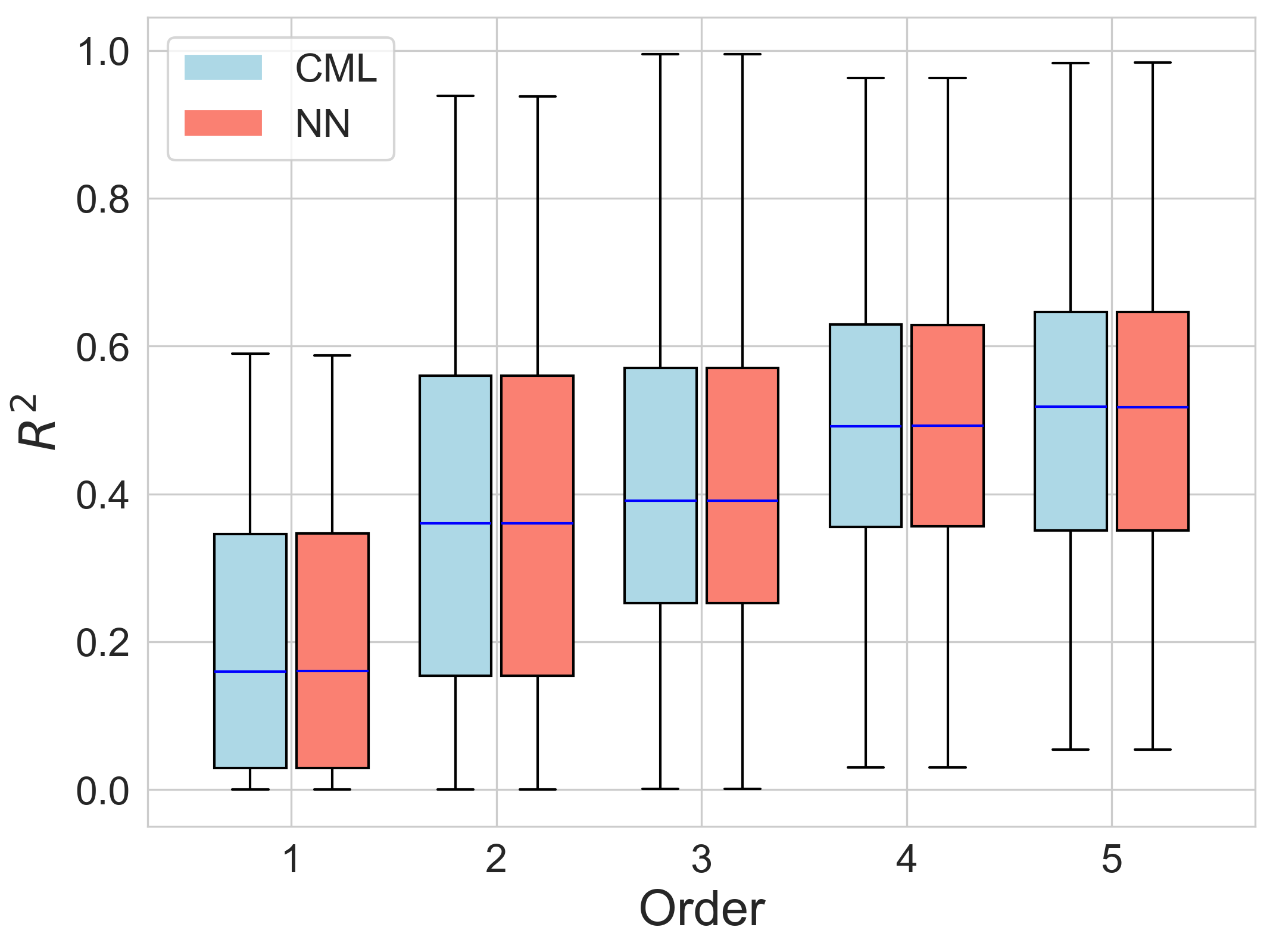} &
\includegraphics[scale=0.3]{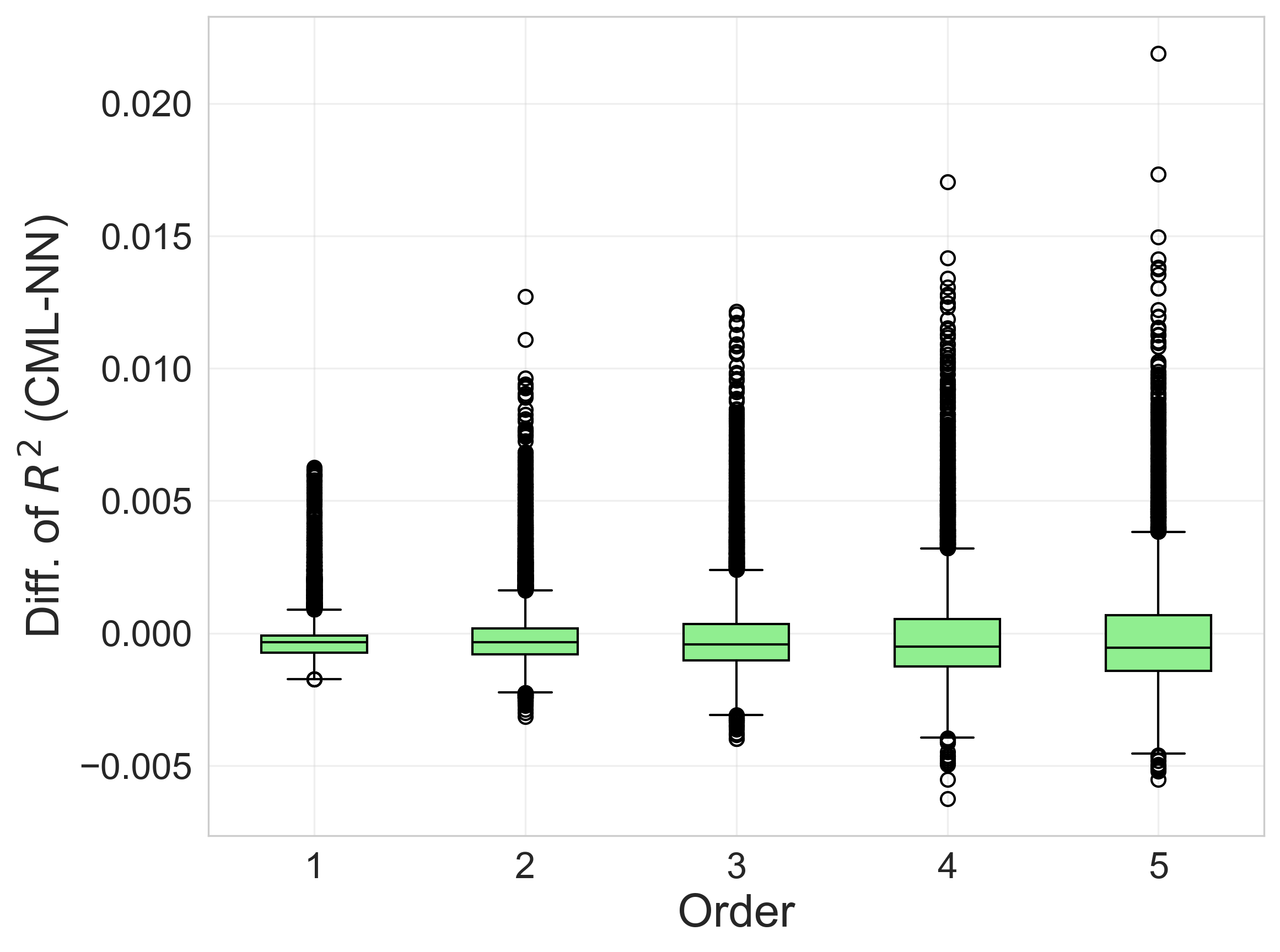} \\
\end{tabular}
\caption{Performance comparison by AR order $p$: (a) computation time ratio, (b) ratio without outliers, (c) $R^2$, and (d) paired $R^2$ differences.}
\label{fig:scaling_results}
\end{figure}

To assess scalability with model order $p$, Fig.~\ref{fig:scaling_results} compares computation time and estimation quality for increasing $p$. Panels (a,b) show the computation time ratio (CML/NN). As $p$ increases, the median ratio grows from 12.6 for $p=1$ to nearly 35 for $p=5$, indicating that the NN estimator scales better and remains consistently faster than CML. Panel (b), which excludes extreme outliers, highlights this trend more clearly.
Panels (c,d) evaluate estimation quality using the coefficient of determination ($R^2$). The $R^2$ distributions for CML and NN are very similar across all orders, reflecting the small differences previously observed in the estimated coefficients. The paired $R^2$ differences (CML$-$NN) have a slightly negative median that decreases with $p$, while variability increases with model order. On average, the difference remains around $-10^{-4}$ for $p=2$ to $p=5$, indicating that NN maintains comparable accuracy while achieving substantially better computational scaling.

\section{Conclusion and Future Work}

This work proposes a neural network formulation that embeds the exact AR($p$) structure in a feedforward network, enabling coefficient estimation through backpropagation while preserving interpretability. The weights are reparameterized using the Durbin–Levinson recursion to ensure stationarity of the estimated process. Numerical simulations on 125,000 synthetic time series of orders $1\leq p \leq 5$ show that CML fails to converge in about 55\% of the cases, particularly when roots approach the unit circle, while the NN estimator successfully recovers coefficients for all time series. When CML converges, both methods achieve very similar coefficient errors, $R^2$, and cost function values. In terms of efficiency, the NN consistently outperforms CML, with a median speedup of 12.6× overall and 34.2× for $p=5$, without loss of estimation accuracy. These results suggest that the proposed approach provides an efficient alternative for AR parameter estimation, preserving the classical statistical interpretation while leveraging gradient-based optimization. Future work will extend this framework to other models of the ARMA class as well as other architectures, such as recurrent NN.


\begin{credits}
\subsubsection{\ackname} 
This work was supported by the Foundation for Science and Technology (FCT) through Institute of Electronics and Informatics Engineering of Aveiro (IEETA) contract doi.org/10.54499/UID/00127/2025. AL acknowledges the individual PhD grant (ref.20/2024/BI/AgendasPRR), funded by Agenda PRR ``NEXUS''.  

\end{credits}
%
%
%
\bibliographystyle{splncs04}
\bibliography{bib}
%

\end{document}